\documentclass[
twocolumn,
]{ceurart}

\begin{document}

\copyrightyear{2023}
\copyrightclause{Copyright for this paper by its authors.
  Use permitted under Creative Commons License Attribution 4.0
  International (CC BY 4.0).}

\conference{CLiC-it 2023: 9th Italian Conference on Computational Linguistics, Nov 30 — Dec 02, 2023, Venice, Italy}

\title{Camoscio: an Italian Instruction-tuned LLaMA}

\author[1]{Andrea Santilli}[%
email=santilli@di.uniroma1.it
]
\address[1]{Sapienza University of Rome - Computer Science Department}

\author[1]{Emanuele Rodolà}[%
email=rodola@di.uniroma1.it
]

\begin{abstract}
In recent years Large Language Models have improved the state of the art on several natural language processing tasks.
However, their availability is frequently restricted to paid API services, posing challenges for researchers in conducting extensive investigations.
On the other hand, while some open-source models have been proposed by the community, they are typically English-centric or multilingual without a specific adaptation for the Italian language.
In an effort to democratize the available and open resources for the Italian language, in this paper we introduce Camoscio: a language model specifically tuned to follow users' prompts in Italian.
Specifically, we finetuned the smallest variant of LLaMA (7b) with LoRA on a corpus of instruction prompts translated to Italian via ChatGPT.
Results indicate that the model's zero-shot performance on various downstream tasks in Italian competes favorably with existing models specifically finetuned for those tasks.
All the artifacts (code, dataset, model) are released to the community at the following url: \href{https://github.com/teelinsan/camoscio}{https://github.com/teelinsan/camoscio}
\end{abstract}


\maketitle

\section{Introduction}

In recent years, Large Language Models (LLMs) have made remarkable advancements in the field of natural language processing, demonstrating state-of-the-art performance on various tasks \cite{brown2020language,chowdhery2022palm,openai2023gpt4}. 
However, the majority of these models are typically controlled by for-profit organizations that release just a paid API for receiving responses based on input textual prompts.
This severely constrains researchers from conducting comprehensive and meaningful research, as they lack access to both the model's weights and the training data regime. 
This limitation is particularly relevant for privacy-sensitive applications (e.g., medical domain) where data cannot be shared with external providers.

On the other hand, several open-source models\footnote{Actual openness depends on the model license.} have been proposed as an alternative to closed models \cite{zhang2022opt,scao2022bloom,touvron2023llama}.
However, most of these models are English-centric or multilingual, albeit with performance that lags behind their monolingual counterparts.
Furthermore, in these latter models, support for the Italian language is usually poor.
For example, BLOOM -- the largest open multilingual model available up to date -- has not been trained on any Italian data, while LLaMA has only a small percentage of training data in the Italian language \footnote{Less than 4.5\% of training data comes from Wikipedia in 20 different languages, including Italian.}. 
In addition to this, most of these models are only trained with the standard language modeling objective (i.e., predict the next token given the previous ones) on corpora of raw textual data, while it has been shown that a second training step of instruction-tuning is crucial to increase downstream performance \cite{sanh2022multitask,wei2021finetuned,chung2022scaling}.
Recently, a step in this direction has been made by \citet{alpaca} with the release of Stanford Alpaca, an instruction-tuned version of LLaMA for the English language. Following this approach, in this paper we propose Camoscio as an instruction-tuned version of LLaMA for the Italian language by translating to Italian the instruction-tuning dataset of Stanford Alpaca.
In particular, we finetuned the smallest version of LLaMA (7 billion parameters) with LoRA \cite{hu2022lora}, a parameter-efficient finetuning technique that allows to train larger models on standard desktop hardware.

Our contributions are the following:
\begin{itemize}
    \item We introduce an instruction-tuning dataset for the Italian language, stemming from the Stanford Alpaca \cite{alpaca} dataset, translating it to Italian.
    \item We train Camoscio on this dataset and evaluate its zero-shot performance on several downstream tasks for the Italian language (NewsSum-IT, SQuAD-IT, XFORMAL IT).
    \item We release all the artifacts (code, dataset, model checkpoints) to the community.
\end{itemize}

\section{Background}

Large language models have emerged as a general class of models capable of performing a wide range of tasks without explicit finetuning by just leveraging in-context examples \cite{bommasani2021opportunities}. They've garnered popularity not only in the natural language processing domain but also across audio, image, and multimodal domains \cite{postolache,dosovitskiy2021an,10.1145/3539618.3591930}, with most of the approaches scaling or optimizing their performance \cite{chowdhery2022palm,santilli-etal-2023-accelerating}.

In the context of the Italian language, the availability of pre-trained language models is currently limited; generic multipurpose LMs are almost nonexistent. 
Notable mentions include: AlBERTo \cite{polignano2019alberto}, an Italian version of BERT \cite{devlin-etal-2019-bert} trained on Italian tweets from TWITA \cite{basile2018long}; GePpeTto \cite{geppetto}, a version of GPT-2 base (117 million parameters) finetuned using Italian Wikipedia and the ItWac corpus \cite{baroni2009wacky}; IT5 \cite{sarti2022it5} a T5 model tailored for Italian using a refined version of the mC4 corpus \cite{xue-etal-2021-mt5}; and BART-IT \cite{fi15010015}, an Italian variant of BART \cite{lewis-etal-2020-bart} trained on the same mixture of data as IT5.
Concurrently to our work, \citet{bacciu2023fauno} proposed Fauno, an Italian version of Baize \cite{xu2023baize} that is a LM trained on a corpus of self-chat performed by ChatGPT. Compared to our work, their approach is tailored to develop a conversational agent for the Italian language. 
After our work, \citet{stambecco} released on their GitHub repository an instruction-tuned version of LLaMA on a translation to Italian of the GPT-4-LLM dataset \cite{peng2023instruction}.

\section{Method}
For the construction of our instruction-tuning dataset for the Italian language, we stem from the Stanford Alpaca dataset \cite{alpaca} and Alpaca LoRA \cite{alpaca-lora} for their finetuning approach.
\subsection{Dataset}
\begin{figure}[t]
\includegraphics[width=6.5cm]{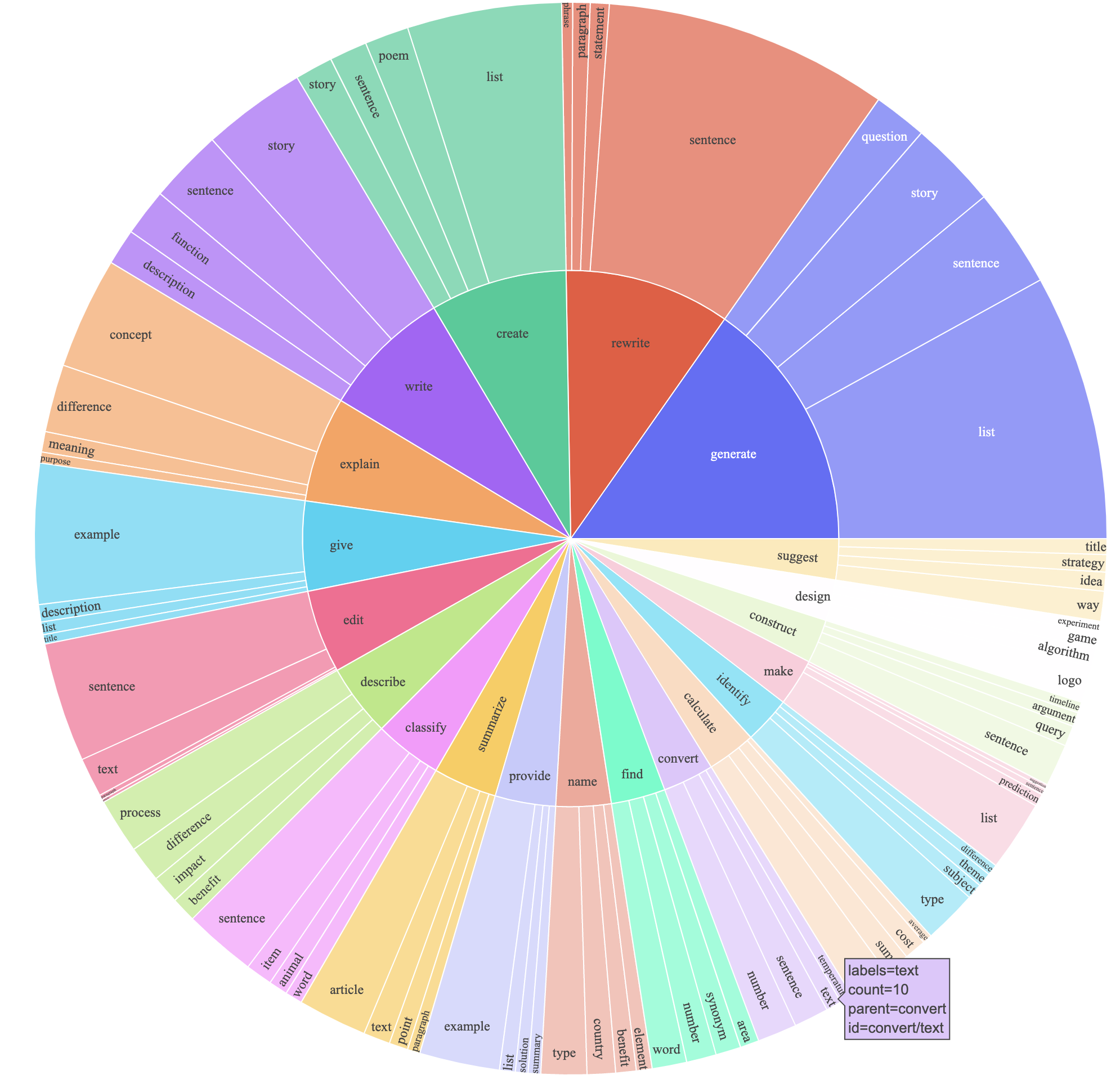}
\centering
\caption{\label{fig:1}Diversity of the examples in the Stanford Alpaca dataset. Illustration from Taori et al. (2023). The inner circle shows the root verb on the instruction while the outer circle shows the direct object. The dataset of Camoscio is constructed by translating all these examples to Italian via \textit{gpt-3.5}.}
\end{figure}

\paragraph{Stanford Alpaca} is an instruction-tuning dataset constructed using the self-instruct method \cite{wang-etal-2023-self-instruct}. Specifically, the authors started with a set of 175 human-written instruction-output pairs from the original self-instruct paper\footnote{\href{https://github.com/yizhongw/self-instruct}{https://github.com/yizhongw/self-instruct}} and used them as in-context examples to prompt OpenAI \textit{text-davinci-003}. A total of 52.000 novel examples are generated with this technique. Each example includes an \textit{instruction}, in natural English language, the answer (\textit{output)}, and optionally an additional context (\textit{input}) for some datapoints (e.g., a short paragraph for question answering).
Figure \ref{fig:1} shows different types of instructions in the dataset.

\paragraph{Translation.} Inspired by \citet{croce-etal-2018-neural}, \citet{scaiella_et_al:2019} and \citet{larcher2023cabrita}, we translated the original dataset of Stanford Alpaca to Italian using \texttt{gpt-3.5-turbo} with the prompt \textit{``Translate the following text to Italian: }\texttt{\{text\}}\textit{"}. We translated all the fields in the dataset (\textit{instruction, input, output}). We decided to use ChatGPT instead of other APIs for translation (e.g., Google Translate,  Microsoft Azure Translator, DeepL) because we found it to be more robust for translating code examples i.e., it translates correctly just the comments in the code and not also the coding lexicon of the programming language.
We provide here an example from the dataset. \texttt{Instruction}: \textit{``Data una parola, costruisci i suoi antonimi."}, \texttt{Input}: \textit{``Luce"}, \texttt{Output}: \textit{	
``Scuro, pesante, denso"}.

Clearly the translation is not always perfect, but it is a fast-and-cheap method to bootstrap a noisy instruction-tuning dataset for the Italian language.

\begin{table*}[t]
\caption{Results on SQuAD-IT. All the models are trained on the SQuAD-IT training set, except for Camoscio which is evaluated in a zero-shot fashion. The additional evaluation metric \textit{Exact Match via ChatGPT} is highlighted in grey. The scores F1 and EM for competitor models are reported from their respective papers.}
\label{tab:qa}
\begin{center}
\begin{small}
\begin{tabular}[t]{l|cc|c|cccc}
\toprule
 & \multicolumn{7}{c}{\textbf{SQuAD-IT}} \\
\cmidrule(lr){2-8}
  & F1 & EM & {EM-GPT} & R1 & R2 & RL & BS\\
 \hline
DrQA-IT \cite{croce-etal-2018-neural}                         &    .659 & .561 & \cellcolor[gray]{.90} - & - & - & - & -  \\
mBERT \cite {croce-etal-2019-deep}                             &  .760 & .650 & \cellcolor[gray]{.90} - & - & - & - & -  \\
BERT\footnotemark[3] \cite{devlin-etal-2019-bert}               &  .753 & .638 & \cellcolor[gray]{.90} - & - & - & - & -  \\
MiniLM \cite{riabi-etal-2021-synthetic}                       &   .720 & .577 & \cellcolor[gray]{.90} - & - & - & - & -  \\
MiniLM$_{\texttt{+st}}$ \cite{riabi-etal-2021-synthetic}      &   .745 & .620 & \cellcolor[gray]{.90} - & - & - & - & -  \\
XLM-R Large$_{\texttt{+st}}$ \cite{riabi-etal-2021-synthetic} &  {.804} & .676 & \cellcolor[gray]{.90} - & - & - & - & -  \\
\hline
mT5 Small \cite{sarti2022it5}  & .660 & .560 & \cellcolor[gray]{.90} .684 & .617 & .347 & .617 & .712  \\
mT5 Base \cite{sarti2022it5}   & .757 & .663 & \cellcolor[gray]{.90} .745 & .709 & .396 & .708 & .770  \\
\hline
IT5 Small \cite{sarti2022it5}   & .716 & .619 & \cellcolor[gray]{.90} .602 & .671 & .372 & .671 & .743  \\
IT5 Base \cite{sarti2022it5}   & .761 & .663 & \cellcolor[gray]{.90} .600 & .712 & .406 & .712 & .770  \\
IT5 Large \cite{sarti2022it5}  & .780 & {.691} & \cellcolor[gray]{.90} .641 & .730 & .412 & .729 & .784  \\
\hline
\rowcolor[gray]{.90} Camoscio-7b \textbf{(0-shot)}&  .270 & {.077} & .576 & .242 & .133 & .241 & .237 \\
\hline
\end{tabular}

\end{small}
\end{center}

\end{table*}

\subsection{Training \& Prompting}
\label{sec:mod-prompt}
We finetuned the smallest version of LLaMA \cite{touvron2023llama} (7 billion) on an instruction-tuning dataset for the Italian language, obtained by translating to Italian the dataset of Stanford Alpaca as described in the paragraph above.

The model is trained with supervision with the standard objective of predicting the next token given the previous ones. The dataset has \texttt{instruction, input, output} fields, but the \texttt{input} is not available for all data points (e.g., open-ended generation). For such cases, we construct the prompt:
 \textit{``Di seguito è riportata un'istruzione che descrive un task. Scrivete una risposta che completi adeguatamente la richiesta. \#\#\# Istruzione:}
\texttt{\{instruction\}} \textit{\#\#\# Risposta:} \texttt{\{output\}}\textit{"}.
If, instead, the datapoint also has an \texttt{input} (e.g., question answering where the input is the contextual paragraph), we construct the prompt: \textit{``Di seguito è riportata un'istruzione che descrive un task, insieme ad un input che fornisce un contesto più ampio. Scrivete una risposta che completi adeguatamente la richiesta.
\#\#\# Istruzione:}
\texttt{\{instruction\}} \textit{\#\#\# Input:} \texttt{\{input\}} \textit{\#\#\# Risposta:} \texttt{\{output\}}\textit{"}.

At inference time, the same prompt is used to generate the answer. Only the text generated after ``[...] \textit{\#\#\# Risposta:"} is used as final output. We sample from the model using \textit{top-p} sampling \cite{topp} with a temperature of 0.2, $p=0.75$, $k=40$, and beam search with 4 beams.

We refer to Appendix \ref{sec:imp_detail} for the additional implementation details.

\section{Experiments}

Currently, there is a very limited availability of datasets for a solid evaluation of the broad capabilities these general-purpose models possess. This is true for English but especially for the Italian language, although the community is moving towards this direction \cite{basile-etal-2023-uinauil}.
To evaluate our model we decided to follow the same evaluation protocol proposed in \citet{sarti2022it5}.
Compared to their approach, we do not perform any training on the downstream tasks, i.e., we perform just the evaluation on the test set in a zero-shot fashion by providing to the model a textual description of the task (e.g., \textit{``Riassumi il seguente articolo"}).
We compared the performance of our model on standard Italian benchmarks for summarization (NewsSum-IT), question answering (SQuAD-IT), and style transfer (XFORMAL IT).

Compared to \citet{sarti2022it5}, we do not include the Wikipedia for Italian Text Summarization (WITS) corpus \cite{casola-lavelli-2021-wits} since Wikipedia is included in the original training corpus of LLaMA \cite{touvron2023llama}. We also omitted the news style transfer task between ``Il Giornale" to ``La Repubblica" (and vice-versa) based on CHANGE-IT \cite{de2020change}, since Camoscio has no concepts of ``Il Giornale" or ``La Repubblica" styles (i.e., it was never exposed during training or finetuning to this kind of articles, although we recognize it might be interesting to analyze this in a few-shot setting).
We describe in the next paragraphs the three datasets used for the evaluation.

\footnotetext[3]{\url{https://huggingface.co/antoniocappiello/bert-base-italian-uncased-squad-it}}

\begin{table*}[t]
\caption{Results on formality style transfer (XFORMAL IT) for the formal-to-informal (F $\rightarrow$ I) and informal-to-formal (I $\rightarrow$ F) directions. Competitors' scores reported from Sarti and Nissim (2022).}
\label{tab:xformal}
\begin{center}
\begin{small}
\begin{tabular}{l|cccc|cccc}
\toprule
& \multicolumn{4}{c}{\textbf{XFORMAL (IT) F $\rightarrow$ I}} & \multicolumn{4}{c}{\textbf{XFORMAL (IT) I $\rightarrow$ F}} \\
\cmidrule(lr){2-5}
\cmidrule(lr){6-9}
  & R1 & R2 & RL & BS & R1 & R2 & RL & BS \\
\midrule
mT5 Small  & .651 & {.450} & .631 & .666 & .638 & .446 & .620 & .684 \\
mT5 Base   & {.653} & .449 & {.632} & {.667} & .661 & .471 & .642 & .712  \\
\hline
IT5 Small   & .650 & {.450} & .631 & .663 & .646 & .451 & .628 & .702  \\
IT5 Base   & .652 & .446 & .632 & .665 & .583 & .403 & .561 & .641 \\
IT5 Large   & .611 & .409 & .586 & .613 & {.663} & {.477} & {.645} & {.714} \\
\hline
\rowcolor[gray]{.90} Camoscio-7b \textbf{(0-shot)}&  .645 & .436 & .623 & .651 & {.622} & {.428} & {.600} & {.667} \\
\hline
\end{tabular}
\end{small}
\end{center}
\end{table*}

\paragraph{News Summarization. } 

We evaluate the news article summarization capabilities of Camoscio using the dataset NewSum-IT proposed by \citet{sarti2022it5}. This dataset is obtained by merging two newspaper sources (``Fanpage.it" and ``Il Post") scraped by the Applied Recognition Technology Laboratory\footnote{\url{https://huggingface.co/ARTeLab}} and available on the Hugging Face Hub \cite{lhoest-etal-2021-datasets}.
We used only the test split for the zero-shot evaluation and asked the model to generate an answer given the \texttt{instruction} \textit{``Dopo aver letto il testo qui sotto, riassumilo adeguatamente."} provided in the textual prompt and the news text provided as \texttt{input} (complete prompt as explained in \S \ref{sec:mod-prompt}).
We use the same evaluation metrics of \citet{sarti2022it5} and report the average across the two newspapers as in their work.

\paragraph{Question Answering.} 
To assess the model performance on extractive question answering, we used the SQuAD-IT dataset \cite{croce-etal-2018-neural}. This dataset is composed of sets of paragraphs, questions, and answers derived from the original SQuAD dataset \cite{rajpurkar-etal-2016-squad} via machine translation and subsequent filtering of problematic instances. As for the previous datasets, we used just the test split for zero-shot evaluation. The model is asked to generate an answer given the \texttt{instruction} 
\textit{``Dopo aver letto il paragrafo qui sotto, rispondi correttamente alla successiva domanda"}. We evaluated the generated answers using the script from \citet{sarti2022it5}. Furthermore, we also used an additional metric ``ChatGPT Exact Match" to better assess the performance. We explain this metric in the following subsection ``Evaluation Metrics".

\paragraph{Formality Style Transfer.}

We assess the style transfer capabilities of Camoscio using the Italian subset of the XFORMAL dataset \cite{briakou-etal-2021-ola}, hereafter referred to as XFORMAL-IT. The dataset consists of forum messages from the GYAFC corpus \cite{rao-tetreault-2018-dear} automatically translated covering several topics (entertainment, music, family, and relationships). The test set is constructed by using crowdworkers via Amazon Mechanical Turk to collect formal-informal pairs directly in Italian. 
The model is evaluated in both style transfer directions (Formal to Informal and Informal to Formal).
We use only the test split for the zero-shot evaluation and ask the model to generate an answer given the \texttt{instruction} \textit{``Dato il seguente testo scritto in modo formale, riscrivilo in modo informale."} and vice versa according to the style transfer direction.

\subsection{Evaluation Metrics}
We use the same evaluation protocol and scripts of \citet{sarti2022it5}. Specifically, for evaluating lexical matches, we rely on the language-independent ROUGE metric proposed by \citet{lin-2004-rouge} in the variants unigram (R1), bigram (R2), and Longest Common Subsequence (RL).
To gauge semantic correspondence, we employ the trained BERTScore metric \cite{zhang2019bertscore} with a widely used BERT model pre-trained on Italian\footnote{dbmdz/bert-base-italian-xxl-uncased} and the same baseline scores as \citet{sarti2022it5}.
Following previous works, for evaluating the Question-Answering task we employ exact-match (EM) and F1-score (F1). 
However, since Camoscio is not trained on the output distribution of the question-answering dataset, these metrics will fail to assess the correctness of the output since the EM will count as zero even with a correct output but different wording.
To account for these variations, we used an approach similar to \citet{zheng2023judging} that leverages an external LM (in our case \textit{gpt-3.5-turbo}) to judge whether the answer provided by a model is correct (1) or not (0) given the question and the ground-truth answer. We refer to this metric as Exact Match via ChatGPT (EM-GPT) and explain it with additional details in Appendix \ref{sec:em-gpt}.

\label{sec:evaluation-metrics}

\subsection{Results and Discussion}

\paragraph{Question Answering.}
Table \ref{tab:qa} shows the results of Camoscio compared to other methods used in the literature. We observe that the metrics commonly used for the task (Exact Match and F1) are very low compared to all the other models. Although this is generally expected since we are comparing trained models with an untrained one, the exact match score is suspiciously low. Looking at the output responses, we noted that Camoscio produces correct but wordy answers (e.g., \textit{``La crisi petrolifera del 1973 è iniziata nell'ottobre 1973."} instead of \textit{``ottobre 1973"}) making the system to perform bad on this score despite the fact that it produces correct answers.
Since all the other systems are trained on the datasets, they are aligned with the expected target distribution and the exact match metric is an effective choice.
Nevertheless, when it comes to the zero-shot configuration in Camoscio, this conventional metric fails to accurately capture the true performance of the task.

To this end, we evaluated the model also with standard evaluation metrics for generative models (R1, R2, RL, BS). However, we also observe in this case low scores despite the fact that a qualitative examination of the provided answers suggests an overall higher quality. 
This is possibly due to the different lengths between the produced answers (long) and the ground truth (short) and reinforces the necessity of developing a more precise metric to accurately gauge task performance.

For this purpose, we used instead the metric \textit{Exact Match via ChatGPT} explained in \S \ref{sec:evaluation-metrics}. This metric shows that the actual zero-shot performance of Camoscio is in line with the other trained models ($.576$) and it is also way higher compared to the original EM metric ($.077$), confirming the need for another type of metric to evaluate the task in the zero-shot setting.
Results also show that the EM-GPT metric of trained models correlates well with the existing EM metric, even though with a little marginal difference.
This suggests that this metric could serve as an approximate estimation of the model's actual performance, although it might be subject to bias according to the model used for estimation.

\begin{table}[t]
\caption{Results on NewSum-IT}
\label{tab:newssum}
\begin{small}
    
\begin{tabular}[b]{l|cccc}
\toprule & \multicolumn{4}{c}{\textbf{NewsSum-IT}} \\
\cmidrule(lr){2-5}
  & R1 & R2 & RL & BS \\
\midrule
mBART Large~\footnotemark$^,$\footnotemark  & {.377} & {.194} & {.291} & - \\
\midrule
mT5 Small   & .323 & .150 & .248 & .375 \\
mT5 Base    & .340 & .161 & .262 & .393 \\
\midrule
IT5 Small   & .330 & .155 & .258 & .386 \\
IT5 Base    & .339 & .160 & .263 & .044 \\
IT5 Large  & .251 & .101 & .195 & .315 \\
\hline
\rowcolor[gray]{.90} Camoscio-7b \textbf{(0-shot)}& .250 & .104 & .174 & .190\\
\hline

\end{tabular}
\end{small}
\end{table}

\footnotetext[6]{\url{https://huggingface.co/ARTeLab/mbart-summarization-ilpost}}
\footnotetext[7]{\url{https://huggingface.co/ARTeLab/mbart-summarization-fanpage}}

\paragraph{Style Transfer \& Summarization.}
Tables~\ref{tab:xformal} and~\ref{tab:newssum} show results respectively for the formality style transfer and news summarization task.
We can observe that the zero-shot performance of Camoscio in both tasks is competitive with trained models. According to the model and training dataset, these latter might achieve slightly better scores at the expense of a less generalist model.
Looking at the qualitative results, we note however that the summarization task on ``Il Post" and ``Fanpage" is affected by some common failure cases.

\paragraph{Failure Cases.}

The most common failure case consists of the model not producing an answer at all after the input prompt ($4.93\%$ of cases on ``Il Post" and $21.16\%$ cases on ``Fanpage"). 
We think that it might be due to the input document of these examples being too long and out of distribution compared to the training documents seen in the instruction-tuning dataset (max length 256 tokens).
This might confuse the model and trigger the generation of the end-of-sentence token.
Other failure cases include the model responding with a copy of the instruction prompt. For example, we found that in $0.75\%$ of the cases in the ``Il Post" split of NewSum-IT, when asked to summarize the article the model responded with \textit{``Questo articolo non è più commentabile. Abbonati al Post per commentare le altre notizie."}. We found this behavior peculiar, considering that the input prompt never mentions that articles were collected from ``Il Post".
Looking carefully at the dataset, we found that these documents were not correctly scraped from the web and the model was simply verbatim copying the input error message above (i.e., the message is already summarized and the model is proposing it as-is).

\paragraph{Open-ended generation.}
We show here some qualitative results on the task of open-ended generation. Compared to the other tasks above, in this task we provide the model with just the \texttt{instruction} without any specific \texttt{input}.
As a tribute to GePpeTto \cite{geppetto}, the first generative language model for Italian, we pose to the model the following \texttt{instruction}: \textit{``Riassumi la storia di Pinocchio"}. The model answered: \textit{``La storia di Pinocchio racconta di un bambino di nome Pinocchio che vive con i suoi genitori in un piccolo villaggio. Un giorno, Pinocchio incontra un burattino magico che gli dona la possibilità di diventare un vero bambino. Pinocchio inizia una serie di avventure mentre cerca di diventare un vero bambino. Durante il suo viaggio, Pinocchio fa amicizia con molti personaggi e impara molte lezioni di vita. Alla fine, Pinocchio raggiunge il suo obiettivo e diventa un vero bambino"}.
At first glance the answer might seem correct, it is grammatically accurate and fluent in Italian. However, knowing the story, the model makes several factual errors like characterizing Pinocchio as a kid (instead of a wooden puppet) living with parents (instead of Geppetto) and introducing a nonexistent magic puppet. Overall this example highlights that, although promising, these models have well-known limitations like hallucinations, factual errors, and several kinds of biases \cite{10.1145/3571730,10.1145/3442188.3445922,sheng-etal-2019-woman}. 
Consequently, it is essential to exercise caution when utilizing them, keeping these limitations in mind.

\section{Conclusion}

In this paper, we introduced Camoscio, a 7 billion instruction-tuned model for the Italian language, together with its Italian instruction-tuning dataset.
Results show that the zero-shot performance of Camoscio on several downstream tasks in Italian is competitive with existing models specifically finetuned for those tasks.
Despite the known limitations of these kinds of models, this is a first step towards a generalist model capable of performing a wide range of tasks in Italian without explicit finetuning.
This is particularly relevant especially in several domains where data is scarce or not available (e.g., medical domain).
In an effort to democratize the available and open resources for the Italian language, we release all the artifacts (code, dataset, model)  to the community.

\section{Limitations}
Results shown in the paper highlight zero-shot performance competitive with existing finetuned models on three different tasks: summarization (NewsSum-IT), question answering (SQuAD-IT), and style transfer (XFORMAL IT). However, it is unclear whether this is true also for other tasks, especially those out of training distribution of the instruction-tuning dataset (see Figure \ref{fig:1}). Evaluating and thoroughly assessing the performance of these kinds of models is still an open research question.
In addition to this, as already mentioned, the model suffers from common problems that affect language models such as hallucinations, factual errors, and several kinds of biases.

\begin{acknowledgments}
We thank Danilo Croce for pointing out existing implementation issues with the tokenization and the training objective in the \textit{alpaca-lora} repository
and Gabriele Sarti for sharing datasets and evaluation protocols used in IT5.
We thank all the anonymous reviewers at \textit{CLiC-it 2023} for the feedback provided.
\end{acknowledgments}

\bibliography{sample-ceur}

\appendix

\section{Implementation Details}
\label{sec:imp_detail}
The model was trained with the LoRA Parameter-efficient Finetuning technique \cite{hu2022lora}, using the Hugging Face Transformers, PEFT, Datasets libraries \cite{transformersLib,peft,lhoest-etal-2021-datasets} and the library Alpaca-LoRA \cite{alpaca-lora}. Specifically, it was trained for 3 epochs with int8 quantization \cite{dettmers2022llm} on a standard desktop GPU Nvidia 3090 on a machine with  Ubuntu 20.04.4 LTS, AMD
Ryzen 9 3900X 12-Core Processor and 32GB of RAM. The model was trained with batches of dimension 4 and gradient accumulation to obtain a final ``virtual batch" of 128. The maximum length used for training is 256 tokens. The learning rate is set to $3\times10^{-4}$ with AdamW \cite{loshchilov2018decoupled} and a total of 100 warmup steps are performed. We used a \textit{lora\_r} (i.e., the dimensionality of the low-rank update of the matrices) equals to 8, \textit{lora\_alpha} equals to 16 and \textit{lora\_dropout} equals to 0.05. We used LoRA adapters just for the matrices \textit{Query} and \textit{Value} in all the attention layers in the LLaMA model, following the original LoRA paper. We used the LLaMA 7 billion checkpoint by loading it from the Hugging Face Hub repository \textit{``decapoda-research/llama-7b-hf"}.

\section{Exact Match via ChatGPT}
\label{sec:em-gpt}
\textit{Exact Match via ChatGPT} is a metric we introduced to evaluate the performance of Camoscio in the zero-shot setting on the question-answering task. This metric assesses whether the answer provided by a model is correct or not, compared to a ground-truth answer, without the need to have an exact string match (Exact Match). Specifically, we used an external LM (in our case \textit{gpt-3.5-turbo}) that acts as a judge with the scope of verifying the correctness of the answer. We used a prompt similar to the following to compute this metric\footnote{Evaluation script available \href{https://github.com/teelinsan/camoscio/blob/main/eval/qa_em_gpt_eval.py}{here}}:

\noindent\\
\textit{``Given the context below and the corresponding question, please indicate whether the answer is correct (1) or not (0). Use a dict format in the response.}\\\\
\textit{Context:} \texttt{\{Context\}}\\
\textit{Question:} \texttt{\{Question\}}\\
\textit{Correct gold answer:} \texttt{\{Correct\_answer\}}\\
\textit{Answer:} \texttt{\{Answer\}}\textit{"}\\

This prompt returns 1 if the answer is correct or zero otherwise. These values are then used to compute the \textit{Exact Match via ChatGPT} with the same procedure of the Exact Match score.

\section{Generation Examples}
\label{sec:add-res}

We propose here some generation examples.
We also provide in the GitHub repository the generated text for all the examples in the test datasets, together with other open-ended generation examples.

\subsection{NewSum-IT (``Il Post")}

\noindent\textit{Input text:}\\
L’esame di maturità è sempre un momento importante, quando lo si vive, anche se poi forse a distanza di anni lo si ricorderà con distacco e pensando che non era poi così difficile, o memorabile. Per molti continuerà a farsi vivo di notte, nei sogni. Quello di quest’anno, che inizia oggi, di certo sarà un po’ diverso: senza tema e dizionari, senza attese di gruppo fuori dalla scuola prima di iniziare, ma con mascherine e gel igienizzanti. Agli studenti della Scuola Holden, la scuola di scrittura di Torino fondata da Alessandro Baricco, è stato chiesto di scrivere una lettera ai maturandi del 2020, per incoraggiarli e dare loro sostegno. Quella scritta da Nausica Anerdi, che ha frequentato il primo anno di Academy, il corso di laurea triennale in scrittura della Holden, è stata pubblicata oggi sul Post e su Repubblica, La Stampa e sulla pagina di Torino del Corriere della Sera. Tutte le altre potete leggerle qui.\\

\noindent\textit{Camoscio generated text:}\\ 
Nausica Anerdi, che ha frequentato il primo anno di Academy, il corso di laurea triennale in scrittura della Holden, ha scritto una lettera ai maturandi del 2020, per incoraggiarli e dare loro sostegno.\\

\noindent\textit{mT5-base generated text:}\\ 
La lettera ai maturandi del 2020. È stata scritta da Nausica Anerdi, che ha frequentato il primo anno di Academy, la scuola di scrittura di Torino fondata da Alessandro Baricco.\\

\noindent\textit{Ground Truth:}\\ 
Una lettera a chi sta per fare la maturità. Da parte di una studentessa della Scuola Holden, del corso di laurea triennale in scrittura.
\rule{\linewidth}{0.4pt}

\vspace{2mm}

\subsection{NewSum-IT (``Fanpage.it")}

\noindent\textit{Input text:}\\
Continuano ad aumentare i nuovi casi di coronavirus nel nostro Paese. Sono stati 2.800 i contagi registrati ieri: numeri che preoccupano il governo e che ricordano quelli delle fasi più critiche dell'emergenza. Domani l'esecutivo si riunirà e valuterà se sia il caso di rendere più severe le norme anti-contagio attualmente in vigore. Entro la prossima settimana si attende il nuovo Dpcm contenente le misure di contrasto all'epidemia, mentre si valuta la proroga dello stato di emergenza fino al prossimo 31 gennaio 2021. Ma vediamo quindi quali sono queste nuove regole che il governo sta pensando di introdurre per frenare la curva dei contagi. L'obbligo di portare la mascherina all'aperto, già introdotto nei giorni scorsi in alcune zone, sarà esteso a tutto il territorio nazionale. Oltre quindi a confermare la necessità di indossare sempre il dispositivo di protezione nei luoghi chiusi, di igienizzare frequentemente le mani e di rispettare le distanze di sicurezza e il divieto di assembramento, il governo studia se rendere alcune misure più stringenti. In particolare, saranno potenziati i controlli nei luoghi della movida o dove è più facile che si vadano a costituire affollamenti. Le operazioni di vigilanza saranno affidate anche ai militari impegnati nel progetto ``Strade secure". Il ministro della Salute, Roberto Speranza, si sarebbe detto favorevole all'estensione dell'obbligo di mascherina all'aperto a tutto il Paese, mantenendosi per quella linea di prudenza sostenuta fin dallo scoppio dell'emergenza. La stessa considerazione, d'altronde, è condivisa anche dalla autorità locali, dal momento che diversi governatori hanno già emesso delle ordinanze che impongono di coprire naso e bocca all'aperto, con multe salate per chi non rispetta le nuove regole. Il governo continua ad escludere la possibilità di tornare a una situazione di lockdown generale. Ma se i contagi continuassero a crescere a questi ritmi, potrebbero essere imposte alcune limitazioni in specifiche zone. In questo senso si sta anche considerando di introdurre un chiusura per bar e locali alle 22 o alle 23: una chiusura anticipata affinché non si ripetano situazioni di movida, come abbiamo visto anche quest'estate, in cui si favorisce la diffusione dell'infezione. All'aperto non sarà inoltre consentito riunire più di mille persone alla volta (ovviamente sempre secondo le regole del distanziamento): questa regola sarà valida sia per le manifestazioni che per gli eventi sportivi. Anche il limite per gli stadi resta a mille spettatori. Negli spazi chiusi, invece, il tetto rimane a 200 persone: questo vale sia per cinema, teatri e sale da concerto, ma anche per le celebrazioni private come i matrimoni. Le discoteche resteranno sicuramente chiuse, nei ristoranti si dovrà continuare a rispettare il distanziamento e anche nei negozi gli ingressi saranno contingentati.\\

\noindent\textit{Camoscio generated text:}\\ 
Il governo sta valutando l'estensione dell'obbligo di mascherina all'aperto a tutto il Paese.\\

\noindent\textit{mT5-base generated text:}\\ 
Il governo sta pensando di introdurre nuove regole anti-contagio per frenare la curva dei contagi. Il provvedimento è stato esteso a tutto il territorio nazionale, mentre si valuta la proroga dello stato di emergenza fino al prossimo 31 gennaio 2021. Vediamo quindi quali sono le nuove regole che il governo sta pensando di introdurre per frenare la curva dei contagi.\\

\noindent\textit{Ground Truth:}\\ 
Con i nuovi casi di coronavirus in continuo aumento, il governo pensa a inasprire le regole anti-contagio per contrastare la diffusione dell’infezione. Si va verso l’obbligo di indossare la mascherina anche all’aperto e la chiusura a bar e ristoranti alle 22 o alle 23. Anche l’esercito sarà impiegato nei controlli.\\
\rule{\linewidth}{0.4pt}

\vspace{2mm}

\subsection{XFORMAL-IT (F$\rightarrow$I)}

\noindent\textit{Input text:}\\
Buongiorno, quella su E-Bay è molto bella.\\

\noindent\textit{Camoscio generated text:}\\ 
Ciao, quella su E-Bay è davvero bella!\\

\noindent\textit{mT5-base generated text:}\\ 
Buongiorno, quella su E-Bay è molto bella.\\

\noindent\textit{Ground Truth:}\\ 
!ciaoooooooquella su e-bay è troppo carina!!\\
\rule{\linewidth}{0.4pt}

\vspace{2mm}

\subsection{XFORMAL-IT (I$\rightarrow$F)}

\noindent\textit{Input text:}\\
no mai , però ora per recuperare vado subito a lavare i piatti!\\

\noindent\textit{Camoscio generated text:}\\ 
È ora per lavare i piatti.\\

\noindent\textit{mT5-base generated text:}\\ 
No, ma ora per recuperare, vado subito a lavare i piatti.\\

\noindent\textit{Ground Truth:}\\ 
Giammai! Ora però per recuperare vado subito a lavare i piatti\\
\rule{\linewidth}{0.4pt}
\vspace{2mm}

\subsection{SQuAD-IT}
\noindent\textit{Input text:}\\
La crisi petrolifera del 1973 iniziò nell' ottobre 1973 quando i membri dell' Organizzazione dei Paesi esportatori di petrolio arabo (OAPEC, composta dai membri arabi dell' OPEC più Egitto e Siria) proclamarono un embargo petrolifero. Alla fine dell' embargo, nel marzo 1974, il prezzo del petrolio era salito da 3 dollari al barile a quasi 12 dollari a livello mondiale; i prezzi americani erano notevolmente più elevati. L' embargo ha causato una crisi petrolifera, o ``shock", con molti effetti a breve e lungo termine sulla politica globale e sull' economia globale. Più tardi fu chiamato il ``primo shock petrolifero", seguito dalla crisi petrolifera del 1979, definita il ``secondo shock petrolifero". Domanda: Quando è stata la seconda crisi petrolifera?\\

\noindent\textit{Camoscio generated text:}\\ 
La seconda crisi petrolifera è stata nel 1979.\\

\noindent\textit{mT5-base generated text:}\\ 
1979\\

\noindent\textit{Ground Truth:}\\ 
1979\\

\end{document}